\documentclass{bmvc2k}

\usepackage{times}
\usepackage{epsfig}
\usepackage{graphicx}
\usepackage{algorithm}
\usepackage[noend]{algpseudocode}
\usepackage{amsmath}
\usepackage{amssymb}
\usepackage{verbatim}
\usepackage{color}
\usepackage{multirow}
\usepackage{pifont}

\title{Multi-class Novelty Detection Using Mix-up Technique}

\addauthor{Supritam Bhattacharjee}{supritamb@iisc.ac.in}{1}
\addauthor{Devraj Mandal}{devrajm@iisc.ac.in}{1}
\addauthor{Soma Biswas}{somabiswas@iisc.ac.in}{1}

\addinstitution{
 Department of Electrical Engineering\\
 Indian Institute of Science\\
 Bangalore, India-560012.
}

\runninghead{Supritam}{Multi-class Novelty Detection Using Mix-up Technique}

\begin{document}

\maketitle

\begin{abstract}
	\color{black}
	Multi-class novelty detection is increasingly becoming an important area of research due to the continuous increase in the number of object categories.
	It tries to answer the pertinent question: given a test sample, should we even try to classify it?
	We propose a novel solution using the concept of mixup technique for novelty detection, termed as  Segregation Network.
	During training, a pair of examples are selected from the training data and an interpolated data point using their convex combination is constructed. 
	We develop a suitable loss function to train our model to predict its constituent classes.
	During testing, each input query is combined with the known class prototypes to generate mixed samples which are then passed through the trained network.
	Our model which is trained to reveal the constituent classes can then be used to determine whether the sample is novel or not.
	The intuition is that if a query comes from a known class and is mixed with the set of known class prototypes, then the prediction of the trained model for the correct class should be high.
	In contrast, for a query from a novel class, the predictions for all the known classes should be low.
	The proposed model is trained using only the available known class data and does not need access to any auxiliary dataset or attributes.
	Extensive experiments on two benchmark datasets, namely Caltech 256 and Stanford Dogs and comparisons with the state-of-the-art algorithms justifies the usefulness of our approach.
	\color{black}	
\end{abstract}

\section{Introduction}
\label{sec:intro}

Deep learning approaches have achieved impressive performance for object recognition and classification task~\cite{alexnet} \cite{VGG} by using large networks trained with millions of data examples.
However, these networks usually work under a closed set assumption, and thus tries to classify each query sample, even if it does not belong to one of the training classes.
For example, a neural network classifier trained to classify fruits, might classify an input from a completely different category, say ``bird'' into one of the fruit classes with high confidence, which is unlikely to happen if a human does the same task.
To make the systems more intelligent and better suited to real-world  applications ~\cite{pimentel2014review} \cite{markou2003novelty}, they should be able to understand  whether the input belongs to one of the trained classes before trying to classify it.

This problem is addressed in recent literature as out-of-distribution detection \cite{devries2018learning}, anomaly detection \cite{zong2018deep}, novelty detection \cite{mohammadi2018conditional} \cite{abati2019latent} and   one-class classification \cite{perera2019ocgan} \cite{sadooghi2018improving}, each having subtle differences between them. 
One class classification rejects all the classes as outliers except the concerned class.
In out-of-distribution paradigm, the algorithm determines whether samples are coming from other data-sets or distribution.
Such algorithms often require the availability or prior knowledge about the out-of-distribution data for its proper functioning.

In this work, we address the multi-class novelty detection task \cite{perera2019deep}, where given a query, the goal is to understand whether it belongs to one of the training classes.
This is very challenging, since the novel data can come from the same data distribution as that of the training data.
Here, we propose a novel framework, termed {\em Segregation Network}, which utilizes the \textit{mixup} technique \cite{tokozume2018between}  for this task.
The network takes as input a pair of data points, and a third interpolated data point which is generated by mixing them together using a variable ratio.
The goal is to determine the constituent classes and the respective proportions by which the two inputs have been mixed to generate the interpolated data point. 
To this end, we design a novel loss function called Constituency loss for our objective. 
Once the network is trained, given an unknown query, we mix it with the known class prototypes in a predefined proportion and pass it through the network. 
Based on the network output, we infer whether the query belongs to the known set of classes or a novel class unknown to the system.
The main contributions of our work are as follows: 
\begin{itemize}
	\item We propose a novelty detection framework, termed as \textit{Segregation Network}, which uses the mixup technique, to detect whether the test query belongs to one of the known classes or to a novel class.
	\item We design a novel loss function called Constituency loss to train the proposed Segregation Network. 
	\item The proposed network can be trained only using the available training data and does not require access to any auxiliary or out-of-distribution dataset as in done in similar methods like \cite{perera2019deep}.
	This is advantageous as the collection of auxiliary data is often difficult, expensive and might be data dependent with respect to the known class set.
	\item During testing, the proposed network compares the unknown query with the set of known class prototypes.
	We also develop an efficient version of our algorithm, without any significant drop in performance, which utilizes the softmax confidence outputs of the pre-trained network.
	\item We perform experiments on two standard benchmark datasets for novelty detection and the results obtained compare favorably with the state-of-the-art method which leverages auxiliary training data.
\end{itemize}

The rest of the paper is organized as follows.
A brief description of the related work in literature is provided in Section~\ref{related_work}.
The proposed approach is discussed in Section~\ref{method} and the experimental evaluation is described in Section~\ref{expts}.
The paper ends with a brief discussion and conclusion.

\section{Related Work}
\label{related_work}

The foundation of this work is based on two threads of machine learning research, namely novelty detection algorithm and mix-up based learning techniques. \newline

\textbf{Novelty Detection:} This problem is an active area of research for detecting outliers in data.
There have been both statistical \cite{stevens1984outliers} \cite{yamanishi2004line} \cite{kim2012robust}  \cite{eskin2000anomaly}, distance based \cite{knorr2000distance} \cite{hautamaki2004outlier} \cite{eskin2002geometric} and deep learning based approaches \cite{pidhorskyi2018generative} \cite{akcay2018ganomaly} \cite{lee2018hierarchical}.
Statistical methods generally focus on trying to fit the distribution of the known data using probability models \cite{clifton2011novelty}. 
Distance based algorithms generally use some transform and then identify novel classes by thresholding the distance with known examples. 
The assumption is that the known class examples will be much closer to the known class representatives, than the unknown ones, in the transformed space.
A relatively recent work in this direction is Kernel-Null Foley-Sammon Transform (KNFST) \cite{bodesheim2013kernel} for multi-class novelty detection.
Here the same class points are projected into a single point in the null space, and during testing, the distance with respect to the class representative is thresholded to get a novelty score. 
This algorithm was improved to handle incremental incoming classes and subsequently update its novelty detector in \cite{liu2017incremental}. 
Deep learning based approaches such as Open-max~\cite{bendale2016towards} tries to fit a Weibull-distribution to determine the novelty. The generative version of this approach was proposed in \cite{ge2017generative}, where unknown samples were generated. 
Several one-class deep learning based novelty detection has been proposed in recent literature \cite{sabokrou2018adversarially} \cite{perera2019ocgan} \cite{sadooghi2018improving}.
The work in~\cite{perera2019deep} designs a novel training paradigm where a reference set is used to learn a set of negative filters that will not be activated for the known category data. To this end they design a novel loss function called membership loss.
Masana {\em et al.}~\cite{masana2018metric} propose a method to improve the feature space by forming discriminative features with contrastive loss for this task.
Out-of-Distribution detection algorithms \cite{devries2018learning} \cite{liang2017enhancing} \cite{vyas2018out} also addresses a similar task, having subtle difference with the problem addressed in this work.
These approaches assume that the \textit{novel} or out of distribution data is strictly outside the data manifold on which the base network is trained.
Thus, novelty detection is in general more challenging than out of distribution detection \cite{perera2019deep}. \\ \\
\textbf{Mixing:} Learning algorithms involving interpolation or mix-up between classes has been recently introduced in the community. 
In one of the early works, Bengio {\em et al.}~\cite{bengio2013better} employed  \textit{mixing} technique to better understand the underlying factor of disentanglement in data. 
Recently, ~\cite{tokozume2018between} proposed to improve the classification task by interpolating between classes. In~\cite{tokozume2017learning}, sound recognition is done by feeding data between class (or mixed up) samples to improve the performance. 
Another work along similar lines is proposed in~\cite{zhang2017mixup}. 
While the mentioned works interpolate in the input space, \cite{berthelot2018understanding} interpolated in the latent space of auto-encoders to generate images with smooth transition from one class to another using adversarial regularizer. 
Several other works tries to interpolate or mix data from different class in the the latent space of auto-encoders \cite{dumoulin2016adversarially} \cite{mathieu2016disentangling}  \cite{ha2017neural} \cite{bowman2015generating}  \cite{mescheder2017adversarial} for different purposes and using various methods. 
In contrast, in our work, we interpolate in the feature space to train our model.

\section{Motivation}
\begin{figure}[t!]
	\begin{center}        
		\includegraphics[width=0.75\textwidth,height=0.32\textheight]{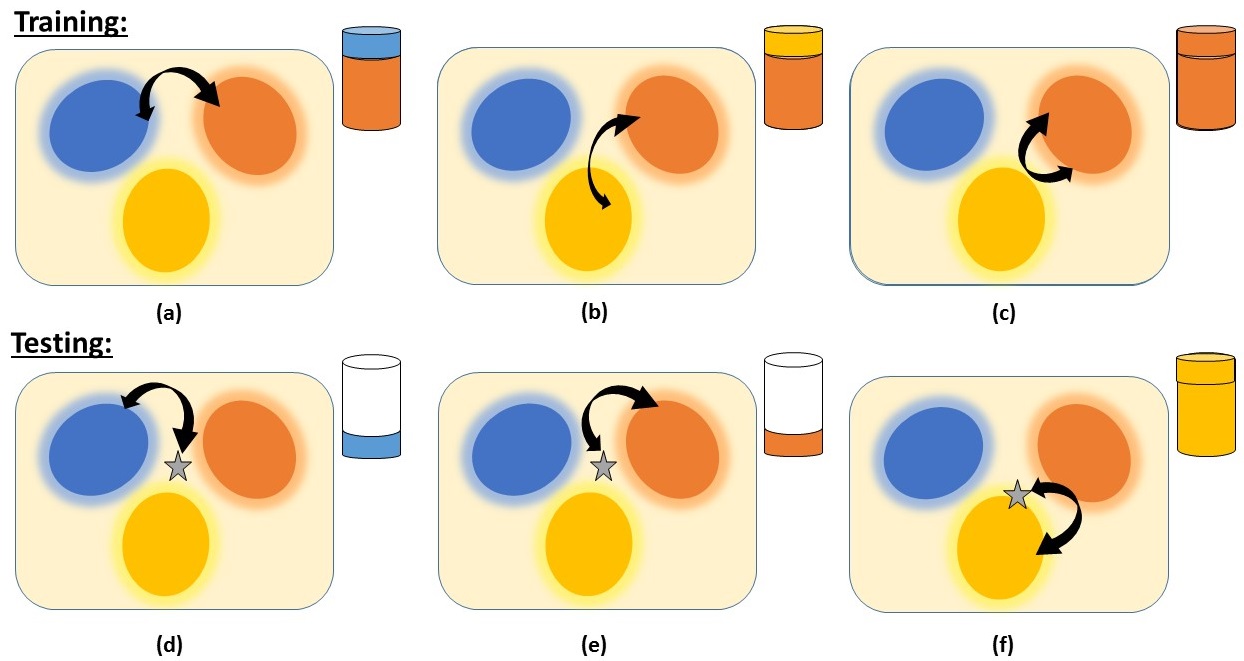}
	\end{center}
	\caption{Illustration showing the motivation behind our proposed approach during the training and testing stage.}
	\label{fig_motivation}    
	\vspace{-15 pt}
\end{figure}

The basic idea behind the development of the algorithm is provided in Figure \ref{fig_motivation}, where we highlight our training (top row) and testing (bottom row) strategies.
The different color schemes indicate the different categories of data.
The bi-directional arrow indicates that samples from these two categories are being used to generate the interpolated data point.
The desired prediction of our trained model is shown in a color-coded bin.

During training, the proposed network aims to segregate the interpolated  point with respect to the two inputs. 
For the example shown in Figure \ref{fig_motivation}(a, b), the interpolated point lies close to the orange class since higher mixing coefficient has been used for that class.
So our model should ideally predict the constituent classes (as shown in the color coded bins) with higher weightage to the orange class.
In Figure \ref{fig_motivation}(c), the interpolated point should be predicted to belong to only the orange class, as two samples from the orange class have been used to generate the interpolated point.

During testing (as shown in bottom row), we generate the interpolated point using the query (denoted as gray star) and the class prototypes of the known classes, and try to predict the constituent classes with respect to the two input points.
We give a larger weightage to the query while generating the interpolated point.
In Figure \ref{fig_motivation}(d,e), since the query does not belong to any of the known class set, the prediction of the network should consist of only the known class mixing coefficient (shown in the color bins).
In case the query comes from the known set (yellow class in Figure \ref{fig_motivation}(f)), the prediction of our network should ideally reflect that with a high value in the color coded bin.
Thus the output of the network can be used to determine whether the input belongs to one of the known classes or to a novel class.

\begin{figure*}[t!]
	\begin{center}        
		\includegraphics[width=0.95\textwidth,height=0.25\textheight]{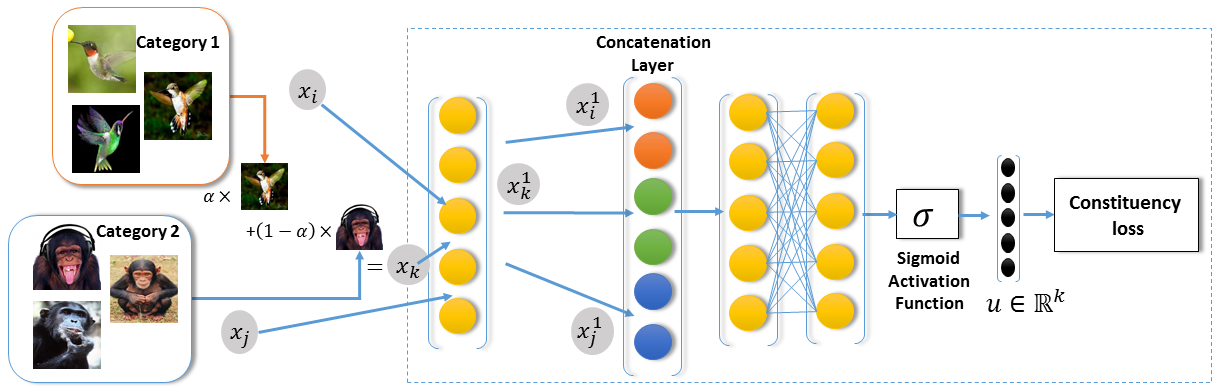}
	\end{center}
	\caption{Illustration of the proposed network. The network accepts feature vectors $x_i$ belonging to $C_1$ (of ``Birds" category) and $x_j$ belonging to $C_2$ (of ``Chimpanzee" category) to create hybrid data in the feature space, $x_k$. All these three vectors $<x_i,x_j,x_k>$ are first passed through the  first fully connected layer of the network to be transformed into a lower dimensional vector before being concatenated together to pass it through the rest of the network. The final activation layer is kept sigmoid so that the output of network $u$ can be used to predict the mixture ratio. We train the model using our proposed novel Constituency loss function.}
	\label{figLC}
	\vspace{-10 pt}
\end{figure*}

\section{Proposed Method}
\label{method}
In this section, we describe the network architecture of the proposed Segregation Network, the novel loss function used to train the model and the training and testing protocol. 
First, we describe the notations used. \\ \\
{\bf Notations:}  Let the input data be represented as $\textbf{X}^{tr} \in \mathbb{R}^{d_t \times N}$, $N$ being the number of training samples and $d_t$ being its feature dimension.
Let the labels be denoted as $\textbf{L}^{tr} \in \mathbb{R}^{K \times N}$, where $K$ is the number of training or known classes. 
We define the known class set to be $C^s = \{C_1, C_2, ..., C_K\}$, and thus $|C^s|=K$.
In the general scenario, the testing data can come from the seen classes or from unseen/novel classes, for which no information is available to the system.
During testing, given a query, the goal is to determine whether it comes from set $C^s$ or not, i.e. whether it belongs to a seen class or a novel class.
Classifying the known examples into its correct class is not the focus of this work and can be done using the base classifier trained using the training data. \\ \\
\textbf{Base Classifier and Features:} Given the training data, a base classifier is trained to classify an input query into one of the given classes.
Since the classifier is now required to work in a general  setting, the input can potentially come from an unseen class, and thus should not be classified by the classifier into one of the seen classes.
Given an input, the novelty detector is designed to take the output features of the classifier (before they are classified) and decide whether the input belongs to one of the seen classes or to a novel class.
If it belongs to a seen class, it can be send back to the classifier for classification.
The proposed novelty detection framework is general and can work with any classifier model.
In this work, we use pre-trained Alexnet~\cite{alexnet} and VGG16 \cite{VGG} architecture as the base classifier. 
These networks are fine-tuned on the respective known class datasets and the extracted features are normalized and given as input to our network.
Now, we describe the details of the Segregation Network.
\subsection{Segregation Network}
The proposed network consists of three fully connected (fc) layers with ReLU activations and dropout between each layer except the final fc layer. 
The final layer is of dimension $K$ (number of seen classes). 
Sigmoid is used as the final layer activation function as the output of Sigmoid is between $[0,1]$ which can be interpreted as the proportion of the mixing classes in our case.
In our design, the network has a $512-1536-256$ architecture, with the numbers denoting the length of each fc layer. We use the Adam optimizer with a learning rate of 0.001 for training our model.
An illustration of the proposed network is shown in Figure~\ref{figLC}.

The network takes as input a triplet set of data samples $\{x_i,x_j,x_k\}$, where $x_i$, $x_j$ are data from the training set and $x_k$ is the mixture obtained by mixing $x_i$ and $x_j$ in some proportion.
Let us denote the output of the first fc layer, which is shared by all three inputs, as $\{x_i^1,x_j^1,x_k^1\}$.
Then $\{x_i^1,x_j^1,x_k^1\}$ is concatenated together to form $x_{ijk}^1$ which is then passed forward through the rest of the network. 
In our implementation, we have used features from the fine-tuned Alexnet \cite{alexnet} or VGG16 \cite{VGG} deep networks, which are of very high dimension.
Thus, the first fc layer serves as a dimensionality reduction layer, the output of which is then concatenated and passed through the rest of the network structure.
The final output of the network after passing through the sigmoid activation function is denoted as $u$. \\ \\
\textbf{Training the model} :
The network is trained such that given an interpolated input, it will decouple/segregate the input data into its constituents.
This property is exploited in the following way.
Given a pair of feature vectors $\{x_i,x_j\}$, we perform  convex combination on this pair to produce $x_k$ , where $x_k = \alpha x_i + (1-\alpha)x_j$, $\alpha \in (0,1)$.
We feed these three feature vectors $\{x_i,x_j,x_k\}$ to our network. 
The output of the network is a $K$ dimensional vector from the final sigmoid layer $u=[0,1]^{K}$.
Since the output is passed through the sigmoid activation function, each element of the $K$-dimensional vector is bounded between $[0,1]$.
In addition, each element denotes the proportion by which the mixed sample $x_k$ has been constructed from that training class. 
For example, an output of $[0,0.6,0.4,0,0]$ (i.e. there are five known classes) indicates that the mixed sample $x_k$ has been constructed as $x_k = 0.6 x_i + 0.4 x_j$ where $x_i \in C_2$ and $x_j \in C_3$.
Given $x_k = \alpha x_i + (1-\alpha)x_j$, the following cases may arise,
\begin{itemize}
	\item If, $x_i \in C_p$ and $x_j \in C_q$, where $p\neq q$, and both $(C_p$, $C_q )\in C^s$ i.e., belongs to the seen classes set, we should get the output of the model such that, $u[p]=\alpha$ and $u[q]=1-\alpha$, while $u[r]=0$ for $r \neq \{p,q\}$. We consider such a pair to be a non-matched pair as the interpolated point $x_k$ lies somewhere in between the two classes based on the value of $\alpha$.
	\item If, both $x_i, x_j \in C_p$, $C_p \in C^s$, the network should ideally output $u[p]=1$ and $u[r]=0$ for $r \neq p$. This is because a mixed element constructed from two data items of the same class should ideally belong to the same class also. We consider such a pair to be a matched pair.
	\item During testing in general scenario, we pair the query sample with class prototypes of the known classes, and so a third case may arise if the query belongs to a novel class. 
	Here, since one of the two inputs to the network is seen, only the output node corresponding to that class should be non-zero and equal to the proportion of this class in the generated mixture.
	We do not explicitly train the network for this scenario, since we do not assume any auxiliary dataset.
	So, we consider only the first two cases for training.
\end{itemize}
Note that as the final activation function is the sigmoid layer and not the standard softmax, total sum of $u$ may not be equal to $1$. 
This is important, since if the input belongs to a novel class, the network will only consider the mixing proportion of the known class. So the proportion of unknown class in the mixture will be ignored and thus the sum will not be equal to 1.

\begin{algorithm}[!t]
	\label{Algo_LC}
	\caption{Algorithm for training the Segregation Network} \label{algo1}
	\begin{algorithmic}[1]
		\State \textbf{Input} : $\textbf{X}^{tr}$ is the input data with their provided labels $\textbf{L}^{tr} \in C^s$.
		\State \textbf{Output} : Trained Segregation Network model to detect novel samples.
		\State \textbf{Initialize} : Initialize the network parameters of Segregation Network. Extract the fine-tuned features for  $\textbf{X}^{tr}$ using Alexnet \cite{alexnet} or VGG16 \cite{VGG}.
		\State Train the network by following these steps
		\State \hspace{3 pt}Randomly generate the mixing coefficient value $\alpha$.
		\State \hspace{3 pt}Randomly take pairs of training data $x_i, x_j \in \textbf{X}^{tr}$ to construct $x_k$ using the mixing coefficient $\alpha$
		\State \hspace{3 pt}Feed-forward the triplet data pair $(x_i,x_j,x_k)$ through the network.
		\State \hspace{3 pt}Compute the constituency loss $\mathcal{L}^{cons}$ and back-propagate it back to train the network.
	\end{algorithmic}
\end{algorithm}

Our network needs to be trained in such a way that the value of $u$ peaks at the position of the constituent classes and also gives the proper mixing proportions.
Since, we do not have the softmax output, we cannot use cross-entropy loss function to train this model. 
In addition, cross-entropy loss function tries to maximize the probability of a sample to belong to a particular class which goes against our desired objective, where a given example can come from outside the known class set. 
Hence, we design a novel loss function termed as \textit{Constituency loss} ($\mathcal{L}^{cons}$) to train our model  which we describe next. \\ \\
\textbf{Constituency loss:} This loss ensures that the output of the Segregation Network, $\{u[r]$, $\quad r=1,...,K\}$ gives positive values for only those classes which has been mixed to create $x_k$.
Thus, the network is expected to output not only the correct proportion over the \textit{mixing} class set $m$, but also give zero output over the \textit{non-mixing} class set $nm$.
Based on this requirement, the  loss function can be written for the $m$ and $nm$ classes as follows
\begin{align}
	\mathcal{L}^{cons} = \sum_{r \in nm} u[r]^2+ g*\sum_{r \in m}(u[r]-\beta[r])^2
	\label{eq:6}
\end{align}
where, $\beta$ denotes the mixing coefficient vector which has zeros for the non-mixing classes, and values of $\alpha$ and $(1-\alpha)$ in their relevant places for the mixing classes.
Let us define $S^0 =\{\beta [r] \in \mathbb{R} \hspace{5 pt} \vert \hspace{5 pt} \beta[r] = 0 \}$ and $S^{\neq 0} =\{\beta[r] \in \mathbb{R} \hspace{5 pt} \vert \hspace{5 pt} \beta[r] \neq 0 \} $ denote the zero element and non-zero element sets. 
The set $S^0$ denotes the sparse set of $\beta [r]$ output for the \textit{non-mixing} classes, while the $S^{\neq 0}$ is for the classes used to form $x_k$.
The weight $g>1$ plays a significant role in training the model as shown in the ablation studies.
This factor is important since  $|S^0| >> |S^{\neq 0}|$, where $|.|$ denotes the cardinality of a set.
Hence during training, we penalize the errors in wrongly predicting the value of $\{\alpha,1-\alpha\}$ much more severely as compared to the incorrect prediction of the zero elements.
In our implementation, we found the best value of $g$ to be between $1000-2000$ in all our experiments.
\subsection{Testing Scenario}
\begin{figure}[h!]
	\begin{center}        
		\includegraphics[width=0.75\textwidth,height=0.20\textheight]{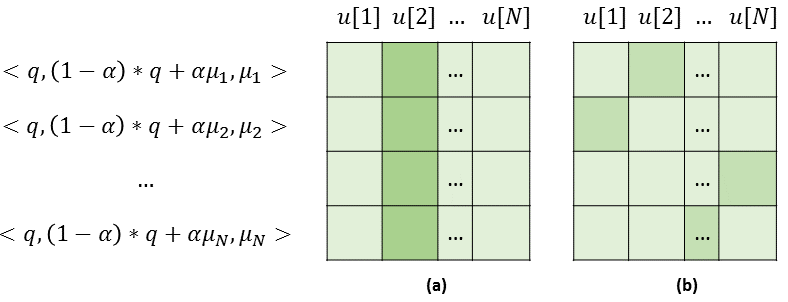}
	\end{center}
	\caption{The procedure to compute the membership score for a given query $q$ is shown here. The set of triplet points $\{q, \mu_r,x_r\} \forall r$ is fed to the network and the prediction is represented as a square matrix. (a) and (b) shows two possible cases when the query $q$ comes from the known and novel set respectively. The deeper shade signifies a higher value than the lighter color in the matrix.}
	\label{algorithm_testing}
	\vspace{-5 pt}
\end{figure}

As mentioned earlier, we assume that a base pre-trained classifier has been fine-tuned on the training classes with a softmax output layer.
Here, it has been taken as the AlexNet \cite{alexnet} or VGG16 \cite{VGG} from where the features are extracted.
In the general testing scenario, the test query can come from one of the seen classes or from a novel class.
Given a test query, we consider the \textit{Top-N} classes which get the highest output scores from the classifier, i.e. the possibility of the query belonging to one of these classes is high.
The goal of the Segregation Network is to consider each of these top classes, and verify whether the query actually belongs to  one of those classes. 
Taking the top few classes is intuitive since (1) if the query belongs to a seen class, its score is usually high and (2) it reduces the computation required for novelty detection using the proposed Segregation Network.
If the query is from a novel class, all retrieved classes are obviously wrong.
Here we use training class centers, $\mu_r$ where $r \in \{1,2,3,..,N\}$, where ($N<K$) as the prototype exemplars.
For each query, $q$, a set of interpolated points is generated as $\{q, \mu_r,x_r\}$, where $x_r=(1-\alpha) q + \alpha \mu_r$, which is then is passed through the proposed network.
The prediction using all exemplar points $\{\mu_r\}_{r=1}^N$ can be viewed as a square matrix (shown in Figure \ref{algorithm_testing}) whose each row signifies the prediction values when paired with a particular cluster center.

The mixing coefficient for the prototype exemplars are kept low while feeding to our model. In other words the mixing coefficient is kept high for the incoming test data. This is because of the following reasons
\begin{itemize}
	\item If the query data belongs to one of the known classes, the high $\alpha$ value for the query example would produce a high $u[r]$ value for the correct known class as shown in Figure \ref{algorithm_testing} (a).
	\item If the query data belongs to an unknown class, the low $(1-\alpha)$ value used for the prototype exemplars forces the network output $u[r]$ to be low for all the (known) classes (as shown in Figure \ref{algorithm_testing} (b)).
\end{itemize}
We define the \textbf{membership score} as the average of the maximum score in each row.
Membership score can be thought of as the likelihood of belonging to the set of known classes.
Thus, for a query coming from the known classes, the membership score should be higher as compared to the membership score of a query coming from the novel class set.
As seen in Figure \ref{algorithm_testing}(a), the individual maximum values of the prediction matrix per row is higher (marked with a deep green shade) and so the membership score (the average value) is higher. When the query belongs to the novel set, as seen in Figure \ref{algorithm_testing}(b), the membership score would come out to be lower.

\section{Experiments}
\label{expts}
In this section, we evaluate our method termed as \textit{Mixing Novelty Detector (MND)} and compare against several state-of-the-art approaches. 
We describe in details the datasets considered for our experiments and the testing protocol.
This is followed by ablation studies to better understand our proposed method.

\subsection{Datasets Used and Testing Protocol}
Here, we report results on two benchmark datasets, namely Caltech 256 \cite{caltech} and Stanford Dogs \cite{dogs}.

\textbf{Caltech 256 dataset \cite{caltech}:} This dataset is a standard dataset for visual recognition consisting of objects spread over $256$ diverse categories.
It consists of $30,607$ images with a minimum of $81$ to a maximum of $827$ image examples per class.
As per the protocol in \cite{perera2019deep}, we took the first $128$ classes as known and rest as unknown.

\textbf{Stanford Dogs dataset \cite{dogs}:} This is a fine grained dataset which has been curated from the ImageNet dataset and consists of images of different breeds of dogs of $120$ categories. 
It consists of total of $20,580$ images.
We consider the first $60$ classes, sorted alphabetically as known. 
The final testing was performed on the remaining $60$ classes, similar to the protocol followed in \cite{perera2019deep}. \\ \\
{\bf State-of-the-art Approaches in Literature:} 
\color{black}
We justify the effectiveness of our proposed model for the task of multi-class novelty detection by comparing against the following current state-of-the-art approaches:
\color{black}
(1) \textbf{Finetune \cite{VGG}:} The fine-tuned network output is taken and thresholded to determine whether a query belongs to the known or novel class;
(2) \textbf{One-class SVM \cite{scholkopf2001estimating}:} All known classes are considered during training the SVM. During testing the maximum SVM score is considered for determining whether a data-point is novel or not.
(3) \textbf{KNFST~\cite{bodesheim2013kernel}:} The deep features are extracted and normalized and the KNFST algorithm is implemented to project all the training samples into the null space, where all examples from a particular class collapse to a single point. Finally, the distance of a query from the training class in the null space is thresholded to determine whether it is novel or not; 
(4) \textbf{Local KNFST \cite{bodesheim2015local}}: This method is similar to that in \cite{bodesheim2013kernel}, but focuses only on the training images most similar to the given query for determining its novelty score; 
(5) \textbf{Openmax \cite{bendale2016towards}:} The feature embedding of the penultimate layer of a trained network is taken and mean activation vectors are determined to fit in the Weibull distribution to finally generate a probability vector of dimension $k+1$, where $k$ is the number of classes.
(6) \textbf{K-extremes \cite{k_extremes}:}  VGG16 features are extracted and the top $0.1$ activation index is used to get the extreme value signatures;
(7) \textbf{Finetune ($c+C$) \cite{perera2019deep}:} The network is trained on additional classes from a reference dataset which has examples other than those present in the main dataset;
and (8) \textbf{Deep Transfer Novelty (DTN):} The state-of-the-art algorithm proposed in \cite{perera2019deep} where an external dataset as reference data is used to learn negative filters which will not get activated for any of the data from the known categories.
The proposed model in \cite{perera2019deep} is trained by using a novel membership loss function. \\ \\
{\bf Evaluation Protocol:} 
\color{black}
We consider area under the receiver operating characteristic curve (AUC) \cite{davis2006relationship} as the evaluation criteria.
AUC is the standard metric used for evaluating these approaches as done in \cite{perera2019deep}.
\color{black}

\subsection{Experimental Results}
\begin{figure}[h!]
	\begin{center}        
		\includegraphics[width=0.95\textwidth,height=0.28\textheight]{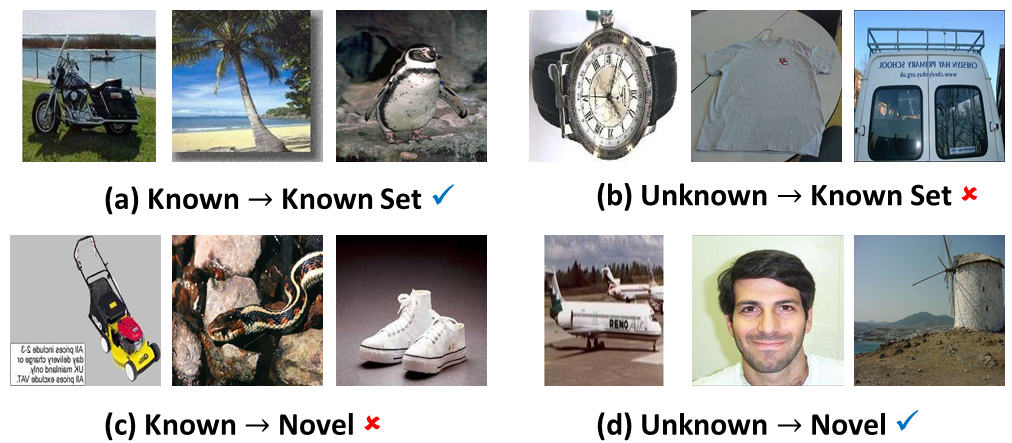}
	\end{center}
	\caption{We display some images from the testing set of the Caltech 256 dataset whose novelty has been determined by our MND algorithm. We show cases where our algorithm has been found to be successful (a,d) and also some unsuccessful cases (b,c). The four cases can be described as (a) known example classified as known, (b) unknown classified as novel, (c) known classified as novel and (d) unknown classified to novel set. }
	\label{classify_examples}
\end{figure}

\color{black}
We evaluate our algorithm by using the features extracted from the pre-trained Alexnet \cite{alexnet} and VGG16 \cite{VGG} networks as done in \cite{perera2019deep}. 
Our results are compared against the current state-of-the-art baseline methods and also evaluated on the same set of features and their results are directly taken from ~\cite{perera2019deep}.
Our results as reported in Table \ref{results} shows that our proposed model gives the best result in three out of four cases.
Our method on VGG16 features has convincingly outperformed the method in \cite{perera2019deep} with a margin of 7.9$\%$ for Stanford Dogs \cite{dogs} and 1.3$\%$ for Caltech 256 \cite{caltech} datasets.
For Alexnet features, it beats all the other approaches for Stanford Dogs and compares favorably for Caltech 256 dataset.
One important point to note is that our proposed framework does not require any external or auxiliary datasets as used in~\cite{perera2019deep} to perform novelty detection.
This makes our approach more suitable for real-world applications.
Since it does not require any auxiliary dataset, it is also computationally lighter than \cite{perera2019deep} as we do not need to train any extra network module.
We also display the novelty detection results on few test example images from the Caltech 256 dataset in Figure \ref{classify_examples}. 
We show successful (a,d) and failure (b,c) cases in Figure \ref{classify_examples}. Successful cases are the ones where a query known (or unknown) sample is classified to belong to the known (or novel) set. 
\color{black}


\begin{table}[h!]
	\centering
	\renewcommand{\arraystretch}{1.0}
	\setlength{\tabcolsep}{2.0 pt}    
	\caption{Comparison of the proposed framework (MND) to the state-of-the-art methods using the AUC of the ROC curve evaluation metric.
		Our algorithm gives convincing results compared to the state-of-the-art Deep Transfer Novelty (DTN), without the use of any extra reference dataset.}
	\label{results}
	\begin{tabular}{|c|c|c|c|c|}
		\hline
		{\bf Dataset} & \multicolumn{2}{c|}{\bf Stanford dogs} & \multicolumn{2}{c|}{\bf Caltech 256} \\ \hline
		& VGG16 & AlexNet & VGG16 & AlexNet \\ \hline \hline
		FineTune\cite{VGG} & 0.766 & 0.702 & 0.827 & 0.785 \\ \hline
		One-Class SVM \cite{scholkopf2001estimating} & 0.542 & 0.520 & 0.576 & 0.561 \\ \hline
		KNFST pre \cite{bodesheim2013kernel} & 0.649 & 0.619 & 0.727 & 0.672 \\ \hline
		KNFST \cite{liu2017incremental} \cite{bodesheim2013kernel} & 0.633 & 0.602 & 0.743 & 0.688 \\ \hline
		Local KNFST pre \cite{bodesheim2015local} & 0.652 & 0.589 & 0.657 & 0.600 \\ \hline
		Local KNFST  \cite{bodesheim2015local} & 0.626 & 0.600 & 0.712 & 0.628 \\ \hline
		K-extremes \cite{bodesheim2015local} & 0.610 & 0.592 & 0.546 & 0.521 \\ \hline
		OpenMax \cite{bendale2016towards} & 0.776 & 0.711 & 0.831 & 0.787 \\ \hline
		Finetune(c+C) \cite{perera2019deep} & 0.780 & 0.692 & 0.848 & 0.788 \\ \hline
		DTN \cite{perera2019deep} & 0.825 & 0.748 & 0.869 & \textbf{0.807} \\ \hline
		\textbf{Proposed MND} & \textbf{0.904} & \textbf{0.773} & \textbf{0.882} & 0.751 \\ \hline
	\end{tabular}
	\vspace{-5 pt}
\end{table}

\subsection{Analysis and Observation}
\begin{figure}[t!]
	\centering
	\begin{center}        
		\includegraphics[width=0.95\textwidth,height=0.20\textheight]{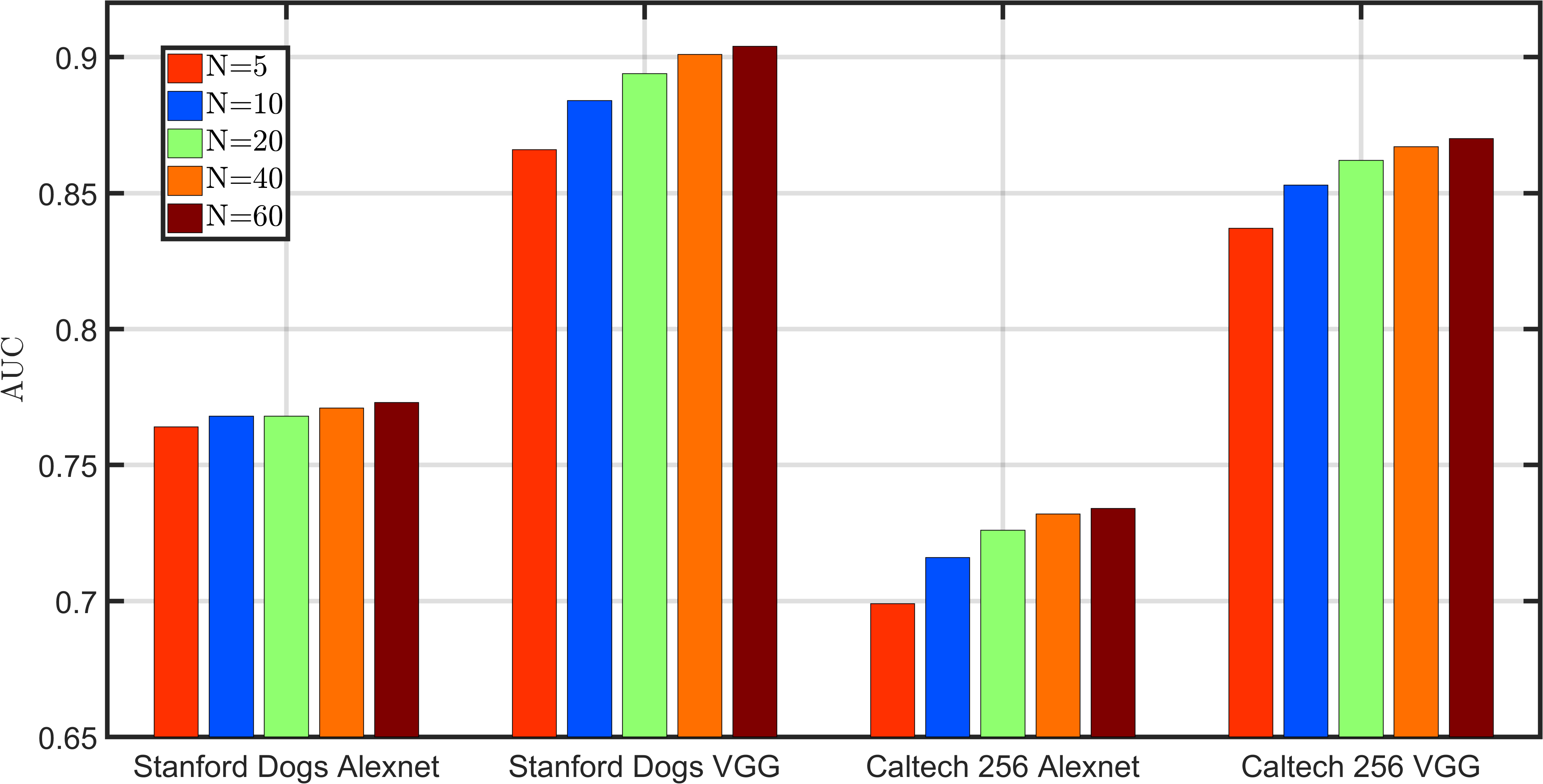}
	\end{center}
	\caption{Performance (AUC) of the proposed MND when the number of class prototypes of known classes ($N$) which is compared with the query is changed. 
		We observe that there is a graceful decrease in AUC when the value of $N$ is slowly decreased for both the datasets using the different feature representations. This naturally leads to a more efficient version of our algorithm.}
	\label{auc_with_prototype}
\end{figure}

We perform extensive analysis to better understand the proposed framework and highlight the salient points. \\ \\
\textbf{Effect of number of top classes used for novelty computation:}
\color{black}
During testing, we compare the query element with all individual class mean  prototypes of the known set for determining whether it is novel or not.
The results are reported in Table \ref{results} when the prototypes of all the known classes ($K$) are used during testing.
Here, we investigate the effect of taking the prototypes of only the top-$N$ classes (as given by the softmax values from the base network, namely Alexnet and VGG16) and provide the results in Figure~\ref{auc_with_prototype}.

\begin{figure}[t!]
	\centering
	\begin{center}        
		\includegraphics[width=0.95\textwidth,height=0.30\textheight]{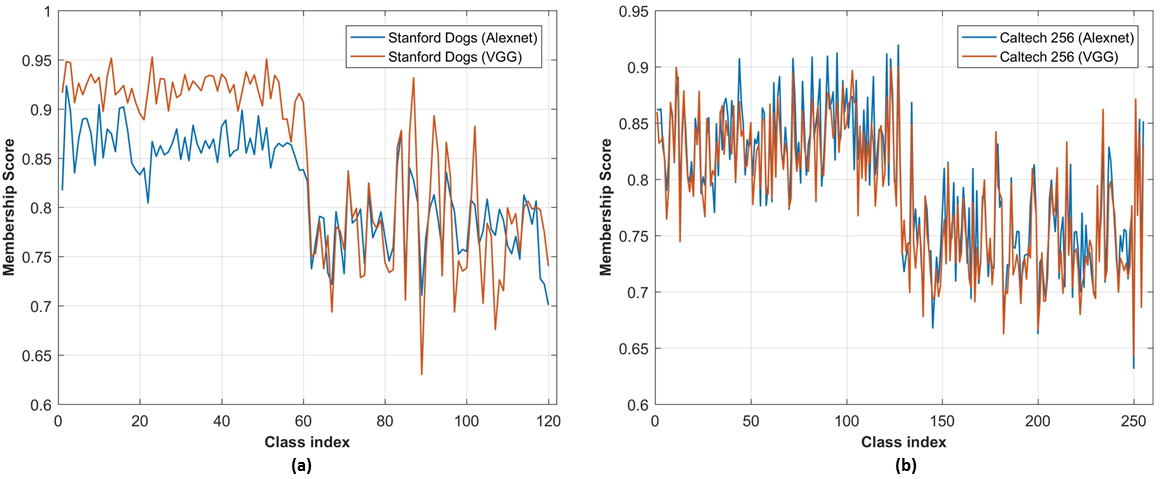}
	\end{center}
	\caption{Average class-wise membership scores for the (a) Stanford Dogs and (b) Caltech 256 dataset using the features of the Alexnet \cite{alexnet} and VGG16 \cite{VGG} network. The first $60$ \& $128$ categories for the two datasets in (a) \& (b) are considered in the known class set while the remaining are considered to form the novel classes. We observed that the class-wise membership score for the known class set in general are higher than that of the novel class examples for both the datasets. This clearly demonstrates the effectiveness of our proposed method.}
	\label{seen_unseen_example}    
	\vspace{-10 pt}
\end{figure}

\begin{figure}[t]
	\begin{center}        
		\includegraphics[width=0.95\textwidth,height=0.35\textheight]{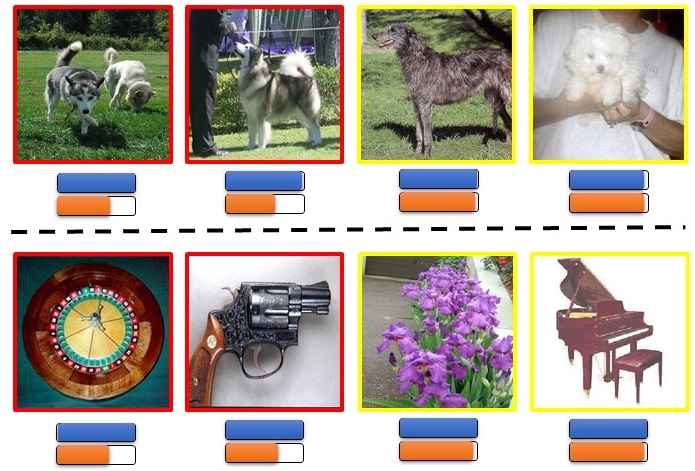}
	\end{center}
	\caption{\color{black} We report the softmax (in blue) and membership score (in orange) for some example test cases from the Stanford Dogs (top row) and Caltech 256 (bottom row). The red (or yellow) bordered images come from the novel (or known) class set. Notice that for a novel query, while softmax values are overly confident, our membership score is low signifying that it is a novel sample. \color{black} }
	\label{softmax_vs_score}    
	\vspace{-5 pt}
\end{figure}

\begin{figure}[h!]
	\begin{center}        
		\includegraphics[width=0.95\textwidth,height=0.25\textheight]{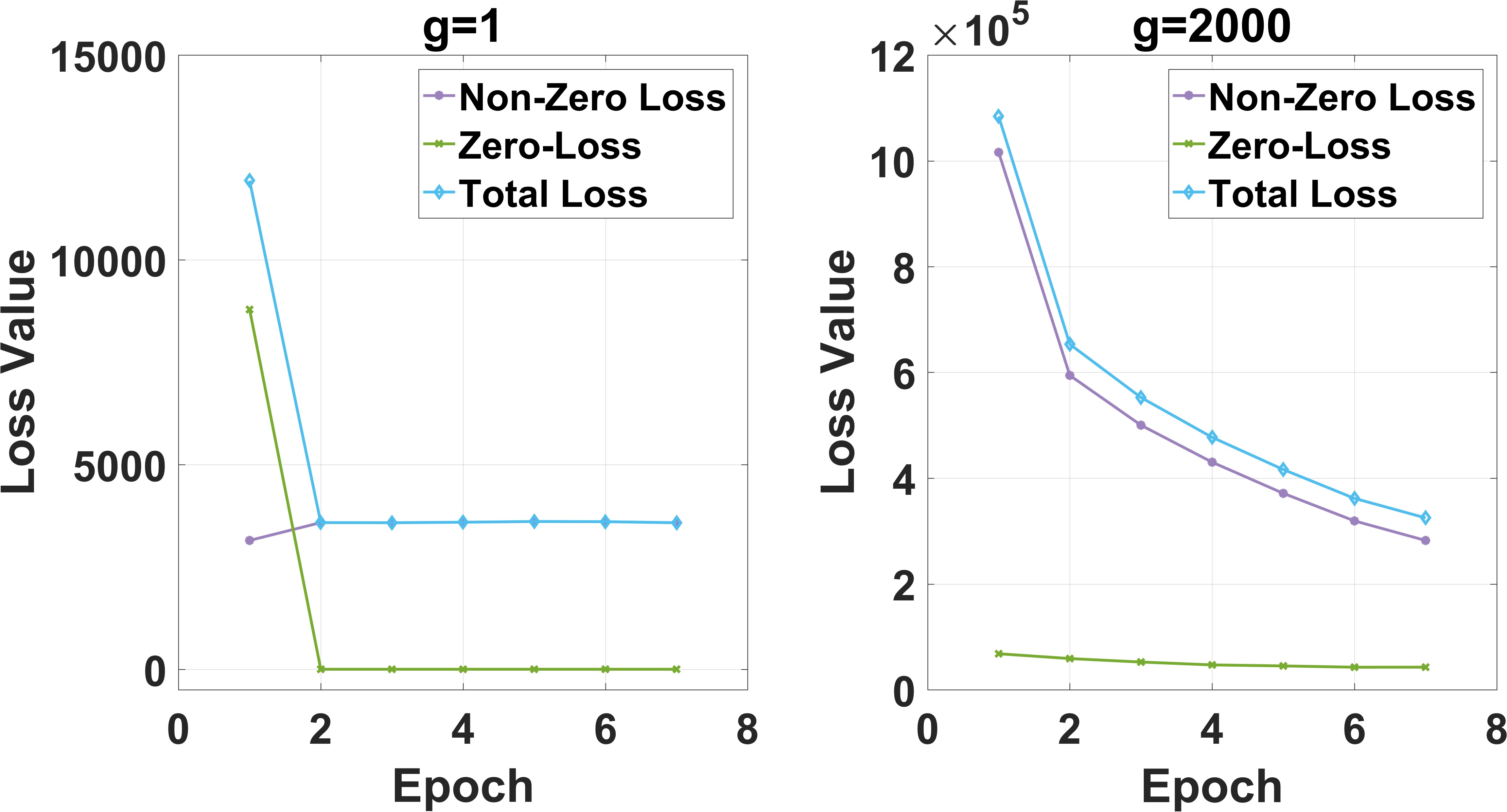}
	\end{center}
	\caption{The plot of the zero-element, non-zero element and total loss with epoch for different values of $g$ in equation \eqref{eq:6} on Stanford Dogs Dataset.}
	\label{g_vary}
	\vspace{-10 pt}
\end{figure}

We observe that there is a very gradual monotonic decrease in the performance of our model as the value of $N$ is decreased from $60$ to $10$.
This helps us to develop a computationally lighter version of our algorithm where we get satisfactory performance even when comparing the query with only the top few class prototypes.
This becomes especially useful when the number of known classes becomes very large.\\ 
\textbf{Analysis of class-wise membership score:}
\color{black}
To better understand why our model is working, we plot the average class-wise membership score for the images of the seen and unseen categories for the (a) Stanford Dogs (first $60$ categories as known) and (b) Caltech 256 dataset (first $128$ categories as known) in Figure \ref{seen_unseen_example}.
Two important conclusions can be drawn from here - (1) the difference in membership score between the seen and novel sets is more pronounced in case of the Stanford Dogs dataset than the Caltech 256 dataset.
This effectively means that our algorithm should perform better for the Stanford Dogs dataset which is further reflected in Table \ref{results}.
(2) The better performance of the VGG16 model over its Alexnet counterpart (as shown in  Table \ref{results}) can be explained due to the fact that the membership profile for VGG16 has higher peaks and lower troughs than Alexnet.
A higher peak and lower trough leads to a better margin for error which results in more ease of determining whether a class is novel or not. \\ 
\textbf{Varying the number of prototypes per class: } 
\color{black}
We have used the mean vector as the class prototype for each class in all our experiments.
We conduct experiments on the Stanford Dogs dataset to analyze here how the proposed framework performs when each known class is instead represented by a set of multiple feature vectors instead of a single mean vector.
The set of representative feature vectors for each class can be determined by applying k-means algorithm on each category and then selecting the subsequent cluster centers which have been formed.
\begin{table}[t!]
	\renewcommand{\arraystretch}{1.00}
	\setlength{\tabcolsep}{8.0 pt}
	\centering
	\caption{ AUC performance  of the proposed method MND with increasing number of prototypes taken for each known class during testing on the Stanford Dogs dataset.}
	\label{no_of_proto1}
	\begin{tabular}{|c|c|c|c|c|}
		\hline
		& \multicolumn{4}{c|}{\textbf{No of prototypes per class}} \\ \hline
		& \textbf{1} & \textbf{2} & \textbf{3} & \textbf{5} \\ \hline
		\textbf{VGG16} & 0.904 & 0.913 & 0.913 & 0.914 \\ \hline
		\textbf{AlexNet} & 0.773 & 0.773 & 0.775 & 0.782 \\ \hline
	\end{tabular}
	\vspace{-2 pt}
\end{table}
Performance in Table \ref{no_of_proto1} show that there is a very slight increase in performance with the increase in the number of prototypes.
We thus select a single prototype to evaluate our model during the testing phase. \\ 
\textbf{Effect of mixing coefficient $\alpha$}:
Our model is trained by randomly sampling the value $\alpha$ ($0 \leq \alpha \leq 1$) and then assigning it to the pair of points.
The purpose of the network is to correctly predict the mixing coefficient.
During testing, we set $\alpha$ to a fixed predefined value.
\begin{figure}[t!]
	\begin{center}        
		\includegraphics[width=0.95\textwidth,height=0.25\textheight]{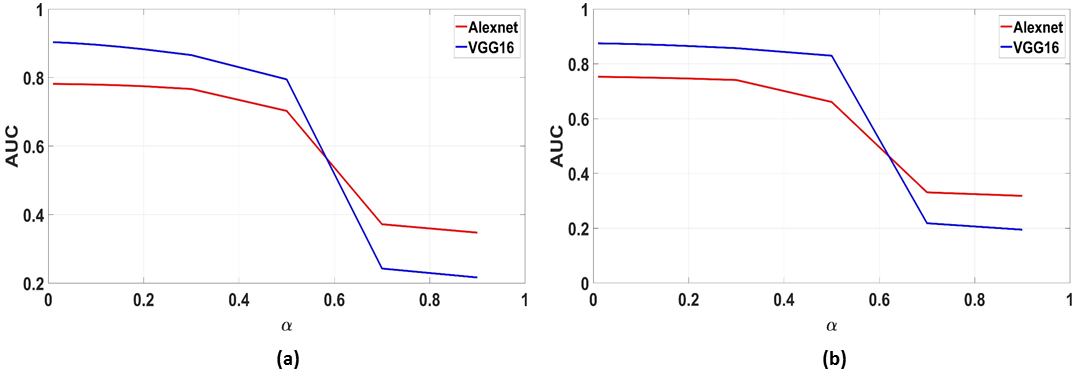}
	\end{center}
	\caption{The effect of $\alpha$ on the performance (AUC) for (a) Standford Dogs and (b) Caltech 256 dataset. Notice that as the value of $\alpha$ is increased the AUC performance drops significantly while the performance saturates at lower values of $\alpha$.}
	\label{alpha_vary_auc}
	\vspace{-10 pt}
\end{figure}
In Figure \ref{alpha_vary_auc}, we report the AUC performance of our model for different values of $\alpha$ and observe that the $\alpha$ value of the cluster center $\mu$ should be kept low to get better performance.
The reason for this being that, the interpolated point $x_k$, formed by $\alpha \mu + (1-\alpha) q$ with, $q$ being the incoming query, and $\mu$ being the class prototype, is much closer to $q$ for a small value of $\alpha$.
Thus if $q$ and $\mu$ belong to the same class then the interpolated point should also belong to the same class leading to the trained model outputting a higher prediction value for that particular class.
Conversely, if $q$ belongs to a novel class altogether, the interpolated point should be far apart from $\mu$ leading the network to give a small value corresponding to that class. \\ 
\textbf{Softmax vs Membership Scores}: We analyze the Softmax score along with its membership value for few test cases in Figure \ref{softmax_vs_score} for the Stanford Dogs and Caltech 256 dataset respectively.
From the figure, it is clear that while the softmax values are overly confident signifying that the image possibly belongs to the known set, our membership score being low helps to determine that it is a novel sample. \\ 
\textbf{Effect of loss weighting factor $g$}:
\color{black}
One of the critical design parameters of our algorithm is the choice of the hyper-parameter $g$ in equation \eqref{eq:6}.
We observe from Figure \ref{g_vary} the training value loss profile for different values of $g$ for Stanford Dogs dataset when using the VGG16 features.
From the figure, we can draw the conclusion that having a weight of $1$ i.e., giving equal importance to the zero and non-zero loss gives poor result as the network only learns to output values close to zero. 
A higher weight assigned to the non-zero element can help to effectively train our model as shown in Figure \ref{g_vary} thereby reducing both the losses.
\color{black}
\section{Conclusions}
\color{black} 
In this paper, we have developed a novel methodology (MND) for multi-class novelty detection which works using the concept of mixup technique.
We have designed a novel constituency loss for training our model.
Experimental evaluations have shown that our model performs favorably with the current state-of-the-art even without having access to any extra auxiliary dataset.
We have also developed an efficient version of our algorithm, without any significant drop in performance, which utilizes the softmax confidence outputs of the pre-trained network.

\bibliography{egbib}
\end{document}